\documentclass[conference]{IEEEtran}
\IEEEoverridecommandlockouts

\usepackage{cite}
\usepackage{amsmath,amssymb,amsfonts}
\usepackage{algorithmic}
\usepackage{graphicx}
\usepackage{textcomp}
\usepackage{xcolor}
\usepackage{lipsum}
\usepackage{booktabs}
\usepackage{jabbrv}
\def\BibTeX{{\rm B\kern-.05em{\sc i\kern-.025em b}\kern-.08em
    T\kern-.1667em\lower.7ex\hbox{E}\kern-.125emX}}
\begin{document}

\title{Neurophysiological Characteristics of Adaptive Reasoning for Creative Problem-Solving Strategy
\thanks{This research was supported by the Institute of Information \& Communications Technology Planning \& Evaluation (IITP) grant, funded by the Korea government (MSIT) (No. RS-2019-II190079, Artificial Intelligence Graduate School Program (Korea University) and  No. IITP-2025-RS-2024-00436857, Information Technology Research Center (ITRC)).}
}

\author{\IEEEauthorblockN{Jun-Young Kim}
\IEEEauthorblockA{\textit{Dept. of Artificial Intelligence}\\
\textit{Korea University}\\
Seoul, Republic of Korea \\
j\_y\_kim@korea.ac.kr}
\\
\IEEEauthorblockN{Gi-Hwan Shin}
\IEEEauthorblockA{\textit{Dept. of Brain and Cognitive Engineering} \\
\textit{Korea University}\\
Seoul, Republic of Korea  \\
gh\_shin@korea.ac.kr}
\and
\IEEEauthorblockN{Young-Seok Kweon}
\IEEEauthorblockA{\textit{Dept. of Brain and Cognitive Engineering} \\
\textit{Korea University}\\
Seoul, Republic of Korea  \\
youngseokkweon@korea.ac.kr}
\\
\IEEEauthorblockN{Seong-Whan Lee}
\IEEEauthorblockA{\textit{Dept. of Artificial Intelligence} \\
\textit{Korea University} \\
Seoul, Republic of Korea \\
sw.lee@korea.ac.kr}
}

\maketitle

\begin{abstract}
Adaptive reasoning enables humans to flexibly adjust inference strategies when environmental rules or contexts change, yet its underlying neural dynamics remain unclear. This study investigated the neurophysiological mechanisms of adaptive reasoning using a card-sorting paradigm combined with electroencephalography and compared human performance with that of a multimodal large language model. Stimulus- and feedback-locked analyses revealed coordinated delta-theta-alpha dynamics: early delta-theta activity reflected exploratory monitoring and rule inference, whereas occipital alpha engagement indicated confirmatory stabilization of attention after successful rule identification. In contrast, the multimodal large language model exhibited only short-term feedback-driven adjustments without hierarchical rule abstraction or genuine adaptive reasoning. These findings identify the neural signatures of human adaptive reasoning and highlight the need for brain-inspired artificial intelligence that incorporates oscillatory feedback coordination for true context-sensitive adaptation.
\end{abstract}
\begin{IEEEkeywords}
electroencephalogram, adaptive reasoning, large language model;
\end{IEEEkeywords}

\section{INTRODUCTION}
\begin{figure*}[htbp]
  \centering
\includegraphics[width=0.85\linewidth,height=\textheight,keepaspectratio]{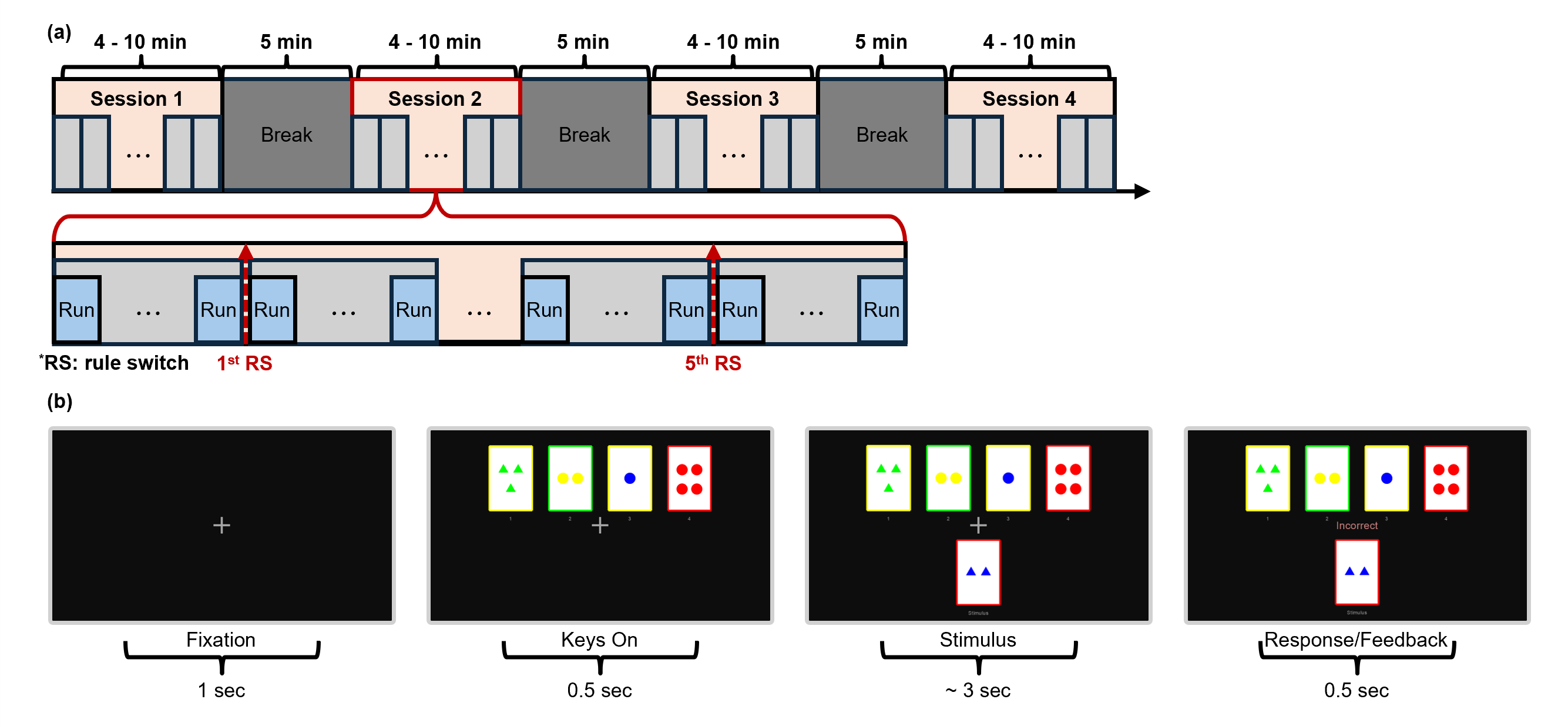}
  \caption{Overview of the WCST-BCI paradigm. (a) Each session consisted of six rule blocks, where four rules (color, shape, number, and border color) were presented in a random order during the first four blocks, followed by two additional blocks in which the rules were randomly selected. The hidden rule automatically switched after ten consecutive correct responses, and short breaks were provided between sessions to reduce fatigue. (b) Within each run, a single trial began with a fixation cross, followed by the presentation of four key cards (“Keys on”) and one stimulus card. Participants selected the matching card according to the current hidden rule, after which feedback (“Correct” or “Incorrect”) was presented on the screen. }
  \label{fig:method}
\end{figure*}
Adaptive reasoning is one of the most distinctive capabilities of the human mind~\cite{Miller2001PFC,Konishi1998Prefrontal}. It enables flexible modification of inference strategies under changing rules or goals through hypothesis updating and knowledge transfer. This ability allows generalization and ambiguity resolution with minimal cues—qualities that remain largely unmatched by current artificial systems. Achieving such energy- and computation-efficient reasoning may require a deeper understanding of the brain mechanisms that realize it~\cite{lee1995multilayer, lee1996multiresolution, Zador2023NeuroAI, lee1999integrated}.

From a cognitive neuroscience perspective, adaptive reasoning emerges from distributed neural population dynamics rather than symbolic computation. The brain forms abstract internal representations of latent variables such as affective meaning or task context~\cite{kim2015abstract}, and electroencephalogram (EEG) captures their rapid temporal dynamics. EEG-based brain-computer interface (BCI) studies show that high-level motor intentions can be classified in real time~\cite{aung2025real, cho2021neurograsp}, that individual cognitive performance can be predicted from spatio-temporal features~\cite{ujma2023multivariate}, and that psychiatric conditions can be discriminated through adaptive optimization frameworks~\cite{prabhakar2020framework}. These works collectively indicate that scalp-recorded neural activity encodes structured, behaviorally relevant information. Such decoding research has historically paralleled advances in neural network modeling, from early recurrent architectures for visual recognition~\cite{lee2003pattern, donahue2015long, lee1997new} to modern deep-learning pipelines for EEG interpretation and behavioral decoding~\cite{lee2015motion}.

Modern multimodal large language models (MLLMs) generate text by probabilistic token prediction. Although contextually coherent, their reasoning is fundamentally statistical rather than inferential~\cite{Webb2023Analogies, malkinski2024reasoning}. Whether such models exhibit true adaptive reasoning—flexibly adjusting internal inference strategies—remains an open question. If they could, they would transcend reactive prediction and achieve context-sensitive, creative, and autonomous intelligence~\cite{Hassabis2017Neuroscience}.

To investigate this, we focus on cognitive flexibility—a hallmark of adaptive reasoning—traditionally assessed by the Wisconsin Card Sorting Test (WCST)~\cite{Nakahara2002SetShifting}. The WCST measures the ability to infer and shift hidden rules from feedback, engaging prefrontal-parietal networks and reinforcement-based control processes~\cite{Botvinick2020DeepRL}. We identify EEG correlates of adaptive reasoning during rule-switching and compare them with MLLM behavioral adaptation, bridging neurophysiological and computational evidence to explore the boundary between biological and artificial reasoning. Recent MLLMs show context-dependent behavior and few-shot adaptability, yet whether they truly perform adaptive reasoning—flexibly altering inference strategies with changing context—remains uncertain, motivating direct comparison with human rule adaptation.


\section{METHODS}

\subsection{Experimental Paradigm and Participants}
We employed a modified WCST-BCI paradigm to examine the neurophysiological basis of adaptive reasoning under dynamic rule-switching conditions. The overall paradigm structure is illustrated in Fig.~\ref{fig:method}(a), and the detailed task display is shown in Fig.~\ref{fig:method}(b). Participants were seated inside a sound-attenuated booth and viewed the task display on a monitor. On each trial, a stimulus card was presented, and participants were required to select one of four target cards using a keyboard within 3~s. The task contained four hidden rules (color, shape, number, and border color), and participants had to infer the current rule through trial-and-error feedback. Once ten consecutive correct responses were achieved, the rule automatically switched, and this process was repeated for six rule blocks per session. Each participant completed four sessions with 5~min rest intervals. Only minimal instructions were given—participants were informed that they must choose by number keys and learn the matching rule based on feedback—to encourage natural adaptation and reasoning in an unfamiliar, dynamically changing environment~\cite{bulthoff2003biologically}.

Five healthy volunteers (mean age = 24.2 ± 0.8 years) participated in the experiment. The study protocol was approved by the Institutional Review Board of Korea University (KUIRB-2023-0429-01), and written informed consent was obtained from all participants prior to the experiment.

\subsection{EEG Acquisition and Preprocessing}
EEG was recorded using a BrainVision Recorder system (Brain Products GmbH, Germany) with an actiCHamp amplifier, slim electrodes, and an actiCAP electrode cap. The data were sampled at 1,000~Hz with 32 channels, with TP9 and TP10 serving as electrooculogram (EOG) channels. The experimental paradigm was implemented using PsychoPy, and both triggers and EEG data were synchronized and collected via the lab streaming layer framework. EEG signals were re-referenced to the common average and notch-filtered at 60~Hz harmonics using spectrum fitting. A band-pass filter of 0.5--100~Hz was then applied. Ocular artifacts were corrected using independent component analysis (ICA) \cite{makeig1995independent, lee2020continuous, subasi2010eeg}: ICA retained components covering 99.9~\% of variance; those correlated with EOG channels ($r>0.4$) were removed. The cleaned data were subsequently band-pass filtered into canonical frequency bands (delta: 0.5--4~Hz, theta: 4--8~Hz, alpha: 8--13~Hz, beta: 13--30~Hz, and gamma: 30--80~Hz) for downstream analyses~\cite{suk2011subject, ding2013changes}.

\subsection{Event-Related Potentials and Topographic Analysis}

Based on prior findings that FRN and P300 reflect feedback evaluation and motivational engagement during decision-making~\cite{Sutton1965P300}, ERPs were extracted to examine their modulation during adaptive reasoning. Epochs were defined from $-0.1$ to $0.5$~s with baseline correction from $-0.1$ to $0$~s. Stimulus-locked trials were classified into confirmation (CONF) and search (SEARCH) phases, and feedback-locked trials into correct (COR) and incorrect (INC) outcomes.

For each participant, condition-wise epochs were averaged with balanced trial counts, and the difference waveform ($\Delta$) between conditions was computed. Group-level effects were tested using cluster-based permutation analysis~\cite{Maris2007Cluster,suk2014predicting} ($p<0.1$, two-tailed), accounting for spatio-temporal dependencies. Significant $\Delta$ clusters within 50~ms windows (0.05-0.50~s) were visualized in topographic maps.


\begin{figure}[tbp]
  \centering
\includegraphics[width=1.0\linewidth,height=\textheight,keepaspectratio]{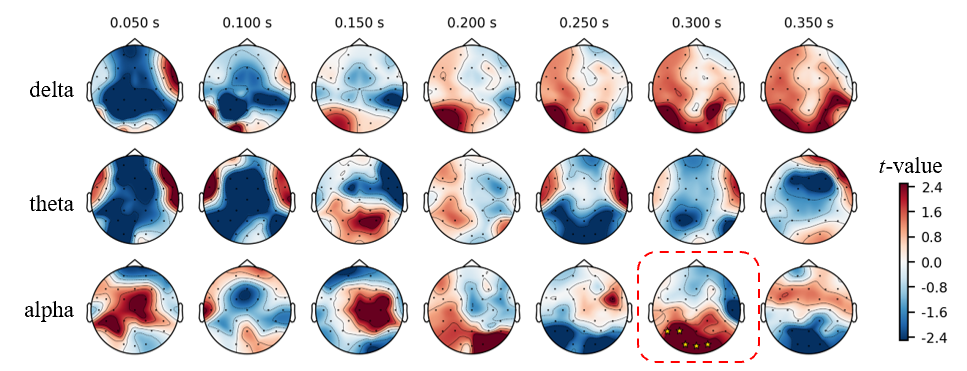}
  \caption{Stimulus-locked band-wise topographic $t$-maps for $\Delta$CONF--SEARCH; red box and yellow dots indicate significant clusters ($p<0.1$).}
  \label{fig:TOPO1}
\end{figure}


\subsection{Multimodal Large Language Model Simulation}
Public pretrained MLLM was evaluated to simulate adaptive reasoning \cite{Brown2020GPT3,Wei2022CoT} within the WCST paradigm. This closed-loop process continued throughout all trials like human. To simulate adaptive reasoning in MLLMs, we evaluated a public pretrained MLLM under the same rule-switching paradigm without gradient updates or fine-tuning, analogous to reinforcement-based adaptive frameworks~\cite{lee2018deep}. On each trial, the model received a single image, which contained the four key cards and one stimulus card, then selected a matching card (1-4) based on in-context feedback from previous trials. The baseline model used was Qwen2.5-VL~\cite{bai2025qwen2}, chosen for its demonstrated few-shot reasoning capability. To quantify behavioral adaptation, we computed (1) rule identification latency—the number of trials required for the model to infer the hidden rule after each switch, (2) accuracy (ACC) within session, and (3) perseverative error rate (PER), defined as consecutive incorrect responses following a rule change. These metrics were compared to human performance to assess whether the MLLM exhibited genuine context-sensitive adaptation rather than statistical pattern repetition.

\section{RESULTS AND DISCUSSION}
\begin{table}[tbp]
    \centering
    \setlength{\tabcolsep}{11pt}
    \caption{Performance of adaptive reasoning for human and multimodal large language model.}
    \label{tab:sfperf}
    \begin{tabular}{rcccc}
    \toprule
     \textbf{Participants}       & \textbf{ACC} $\uparrow$ & \textbf{PER} $\uparrow$  & \textbf{\#RC} $\uparrow$    & \textbf{Latency} $\downarrow$   \\
    \midrule
    \textbf{P1}                  & 72.9        & 0.28         & 5  & 8.37          \\
    \textbf{P2}                  & 73.0        & 0.40         & 5  & 5.09           \\
    \textbf{P3}                  & 81.2        & \textbf{0.45}& 5  & 3.87           \\
    \textbf{P4}                  & 63.6        & 0.39         & 5  & 10.80           \\
    \textbf{P5}                  & \textbf{81.3} & 0.44        & 5  & \textbf{3.78}           \\
    \midrule
    \textbf{MLLMs}              & \textbf{ACC} $\uparrow$ & \textbf{PER} $\uparrow$  & \textbf{\#RC} $\uparrow$    & \textbf{Latency} $\downarrow$   \\
    \midrule
    \textbf{Qwen2.5-VL}          & 21.1        & -    & 0             & 128           \\
    \bottomrule
    \multicolumn{5}{l}{Bold indicates the best value in each column and dataset. }\\
    \multicolumn{5}{l}{RC means rule changed. }
    \end{tabular}
\end{table}

\subsection{Spatiotemporal EEG Analysis}
\subsubsection{Topographic EEG Analysis}
\begin{figure}[tbp]
  \centering
\includegraphics[width=0.98\linewidth,height=\textheight,keepaspectratio]{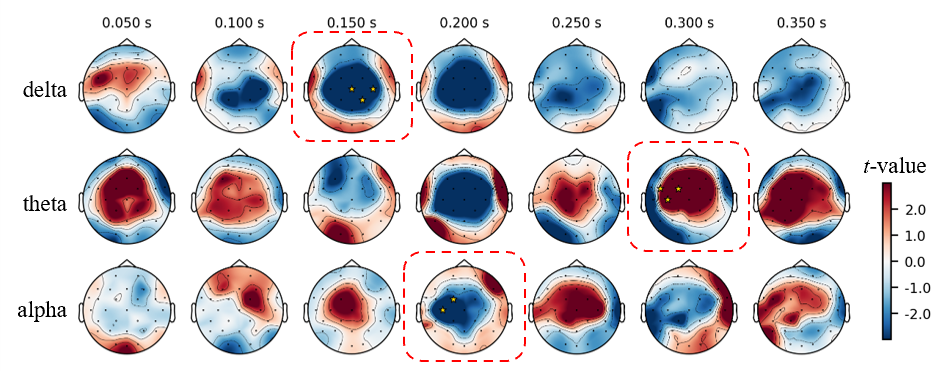}
  \caption{Feedback-locked band-wise topographic $t$-maps for $\Delta$COR--INC; red boxes and yellow dots indicate significant clusters ($p<0.1$).}
  \label{fig:TOPO2}
\end{figure}

Stimulus-locked analyses revealed distinct dominance patterns between the CONF and SEARCH phases (Fig.~\ref{fig:TOPO1}). Early delta-theta activity showed SEARCH dominance over fronto-central regions, indicating enhanced monitoring and exploration during rule inference. Around 300~ms, occipital alpha activity became CONF-dominant, reflecting stabilized attention and sensory integration after successful rule identification.
Feedback-locked responses (Fig.~\ref{fig:TOPO2}) showed dynamic dominance shifts across conditions. Early theta synchronization was stronger for COR trials over parietal regions, while at 150-200~ms, INC trials exhibited broader dominance across theta, alpha, and delta bands, consistent with FRN and error-driven updating~\cite{Holroyd2002FRN}. In later periods, theta activity again became COR-dominant, suggesting re-stabilization of control after correct feedback.
These results indicate that adaptive reasoning alternates between exploratory (SEARCH/INC) and confirmatory (CONF/COR) neural states, coordinated through delta-theta-alpha interactions. For statistical evaluation, a significance threshold ($p<0.1$) was applied due to the limited number of participants.

\subsubsection{Event-Related Potential Analysis}
\begin{figure}[tbp]
  \centering
\includegraphics[width=1.00\linewidth,height=\textheight,keepaspectratio]{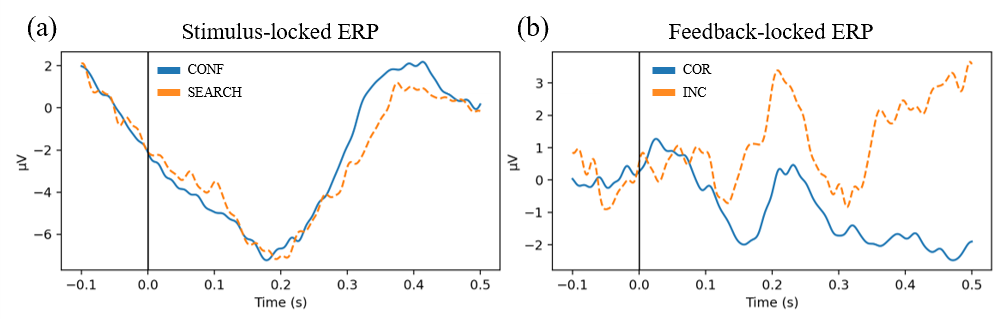}
  \caption{Grand-averaged ERPs at the Cz channel from a representative participant. (a) Stimulus-locked responses for CONF and SEARCH conditions. (b) Feedback-locked responses for COR and INC conditions.}
  \label{fig:ERP}
\end{figure}

Fig.~\ref{fig:ERP} illustrates representative ERPs at the Cz channel. In the stimulus-locked responses, both CONF and SEARCH conditions exhibited a pronounced negative deflection around 200~ms followed by a positive rebound, indicating sequential sensory encoding and cognitive evaluation. The SEARCH phase showed slightly delayed and attenuated positivity, suggesting greater uncertainty during rule inference. In contrast, feedback-locked responses revealed a clear dissociation between COR and INC trials: INC feedback elicited a larger FRN around 200~ms and a subsequent reduced P300 amplitude, reflecting error detection and adaptive rule updating. These ERP dynamics suggest that adaptive reasoning involves coordinated neural processes for prediction error monitoring and contextual adjustment.

\subsection{MLLM Behavioral Analysis}
Compared to human participants (mean accuracy: 74.4\% and rule-switch latency: 6.38 trials), the MLLMs showed markedly lower performance, with an average accuracy of 21.1\% and a latency of 128 trials per rule switch (Table.~\ref{tab:sfperf}). This indicates a fundamental limitation in the MLLMs’ capacity for rule searching and hypothesis testing. Without chain-of-thought reasoning, the models rely on surface-level visual correlations rather than forming and validating internal rules. Consequently, they fail to adapt after incorrect responses and cannot converge effectively. This poor performance likely reflects the absence of working-memory-like processes, feedback-driven learning, and context persistence that humans employ during rule switching.

In contrast, humans can rapidly identify rules with only a few trials, choose responses accordingly, and efficiently explore new rules when rules change. This suggests that the human brain possesses a unique capability absent in current MLLMs, emphasizing the need to design new architectures that mimics this mechanism in human brain. Furthermore, comprehensive evaluation across both short and long chain-of-thought settings, as well as diverse model types, is necessary to fully understand and improve these limitations.

\section{CONCLUSIONS}


Our analyses revealed that adaptive reasoning relies on coordinated delta-theta-alpha dynamics balancing exploratory monitoring and confirmatory stabilization across stimulus and feedback phases. Early delta-theta supports rule inference and uncertainty tracking, while occipital alpha reflects stabilized attention after rule identification. Feedback-related alternation between INC and COR trials highlights dynamic control adjustment for error correction and reinforcement.  
In contrast, MLLMs exhibited only transient feedback adaptation without hierarchical rule abstraction or sustained contextual integration, underscoring their lack of genuine adaptive reasoning mechanisms observed in human cognition.


\bibliographystyle{jabbrv_IEEEtran}
\bibliography{REFERENCE}

\vspace{12pt}
\end{document}